\documentclass[times,review,10pt]{elsarticle}

\myfooter[L]{Author version following major revision.}

\usepackage[letterpaper,top=4.3cm,right=4.8cm,bottom=4.3cm,left=4.8cm]{geometry}
\usepackage{svg}
\usepackage[T1]{fontenc}
\usepackage[utf8]{inputenc}
\usepackage{amsmath,amssymb}
\usepackage{graphicx}
\usepackage{booktabs}
\usepackage{multirow}
\usepackage{pifont}
\usepackage{array}
\usepackage{makecell}
\usepackage{tabularx}
\usepackage{xcolor}
\usepackage{url}
\usepackage{flafter}
\usepackage{placeins}
\usepackage{etoolbox}
\usepackage[font=footnotesize,labelfont=bf]{caption}
\usepackage[hidelinks]{hyperref}
\hypersetup{pdfsubject={Author version following major revision}}

\biboptions{numbers,sort&compress}
\graphicspath{{./}{figures/}}

\newcommand{\prWideTableSetup}{\footnotesize\setlength{\tabcolsep}{1.5pt}\renewcommand{\arraystretch}{1.12}}
\newcolumntype{Y}{>{\centering\arraybackslash}X}
\makeatletter
\patchcmd{\pprintMaketitle}{\Large\@title}{\fontsize{14pt}{17pt}\selectfont\@title}{}{}
\patchcmd{\pprintMaketitle}{\normalsize\elsauthors}{\footnotesize\elsauthors}{}{}
\patchcmd{\MaketitleBox}{\Large\@title}{\fontsize{14pt}{17pt}\selectfont\@title}{}{}
\patchcmd{\MaketitleBox}{\normalsize\elsauthors}{\footnotesize\elsauthors}{}{}
\makeatother

\begin{document}

\begin{frontmatter}

\title{Cross-Scale Sparse Attention for Referring Segmentation in Remote Sensing}

\author[inst1,inst2]{Yuxi Li\fnref{fn1}}
\ead{uniluxli@qq.com}

\author[inst3]{Lu Si\fnref{fn1}}
\ead{silu@tsinghua.edu.cn}

\author[inst4]{Xia Liao\corref{cor1}}
\ead{liaoxia@bupt.edu.cn}

\author[inst4]{Yidong Shen}
\ead{carlshen114514@gmail.com}

\author[inst4]{Zheng Wei}
\ead{wz126543@gmail.com}

\author[inst5]{Jun Zhang}
\ead{jun-zhang@mail.tsinghua.edu.cn}

\author[inst1,inst2]{Xin Yuan}
\ead{xyuan@westlake.edu.cn}

\fntext[fn1]{These authors contributed equally to this work.}
\cortext[cor1]{Corresponding author.}

\affiliation[inst1]{
  organization={Westlake Institute for Advanced Study},
  city={Hangzhou},
  state={Zhejiang},
  country={China}
}

\affiliation[inst2]{
  organization={School of Engineering, Westlake University},
  city={Hangzhou},
  state={Zhejiang},
  country={China}
}

\affiliation[inst3]{
  organization={Beijing National Research Center for Information Science and Technology (BNRist), Tsinghua University},
  city={Beijing},
  country={China}
}

\affiliation[inst4]{
  organization={School of Computer Science, Beijing University of Posts and Telecommunications},
  city={Beijing},
  country={China}
}

\affiliation[inst5]{
  organization={Department of Automation, Tsinghua University},
  city={Beijing},
  country={China}
}

%
%
%
%
%

\begin{abstract}
High-resolution remote sensing (RS) imagery provides fine spatial detail over large areas, making it essential for fine-grained scene understanding. Referring Remote Sensing Image Segmentation (RRSIS) aims to segment regions described by natural language, yet many existing methods rely on sequential alignment--fusion pipelines with limited joint modeling of cross-scale visual hierarchies and linguistic dependencies. Moreover, dense attention mechanisms struggle with the sparsity and wide spatial dispersion of RS targets, while current datasets offer limited structured evaluation of multi-target references, multi-hop relational reasoning, and multi-granularity descriptions, limiting both model development and evaluation. To address these challenges, we propose a Cross-Scale Sparse Attention Model (CSSAM), which aligns multiscale visual features with complex referring expressions through two key modules. The Cross-scale Modality-aligned Fusion Module (CMFM) performs unified cross-scale cross-modal fusion, while the Dynamic Multimodal Sparse Hybrid Attention (DMSHA) selectively attends to informative visual--textual tokens, reducing dense visual interactions. Furthermore, we introduce M$^{3}$RRSIS-D, a RRSIS dataset explicitly designed around three core reasoning dimensions: multi-target (M$^{1}$), reference-dependent multi-hop (M$^{2}$), and multi-granularity (M$^{3}$) expressions. Extensive experiments on M$^{3}$RRSIS-D and two public RRSIS benchmarks demonstrate that CSSAM achieves competitive performance across accuracy, fine-grained reasoning, and high-IoU localization metrics, validating its effectiveness for complex semantic reasoning in realistic RS scenarios. Code, dataset, and model weights will be released upon acceptance.
\end{abstract}


\end{frontmatter}

\section{Introduction}
\label{sec:1}

Large-scale remote sensing (RS) imagery, empowered by recent advances in imaging technology such as array cameras \cite{1,2,3} and high-resolution electro-optical sensors \cite{4}, can now provide high-resolution detail information while covering vast areas up to kilometer-scale \cite{5}. These characteristics make RS imagery well-suited for fine-grained scene interpretation. In particular, Referring Remote Sensing Image Segmentation (RRSIS)---which segments regions described by natural language---has emerged as a core task for aligning visual content with linguistic expressions at the pixel level. It allows a user to select particular objects, such as buildings beside a road, rather than segmenting every instance of a predefined class.

\begin{figure}[!htbp]
  \centering
  \includegraphics[width=\textwidth]{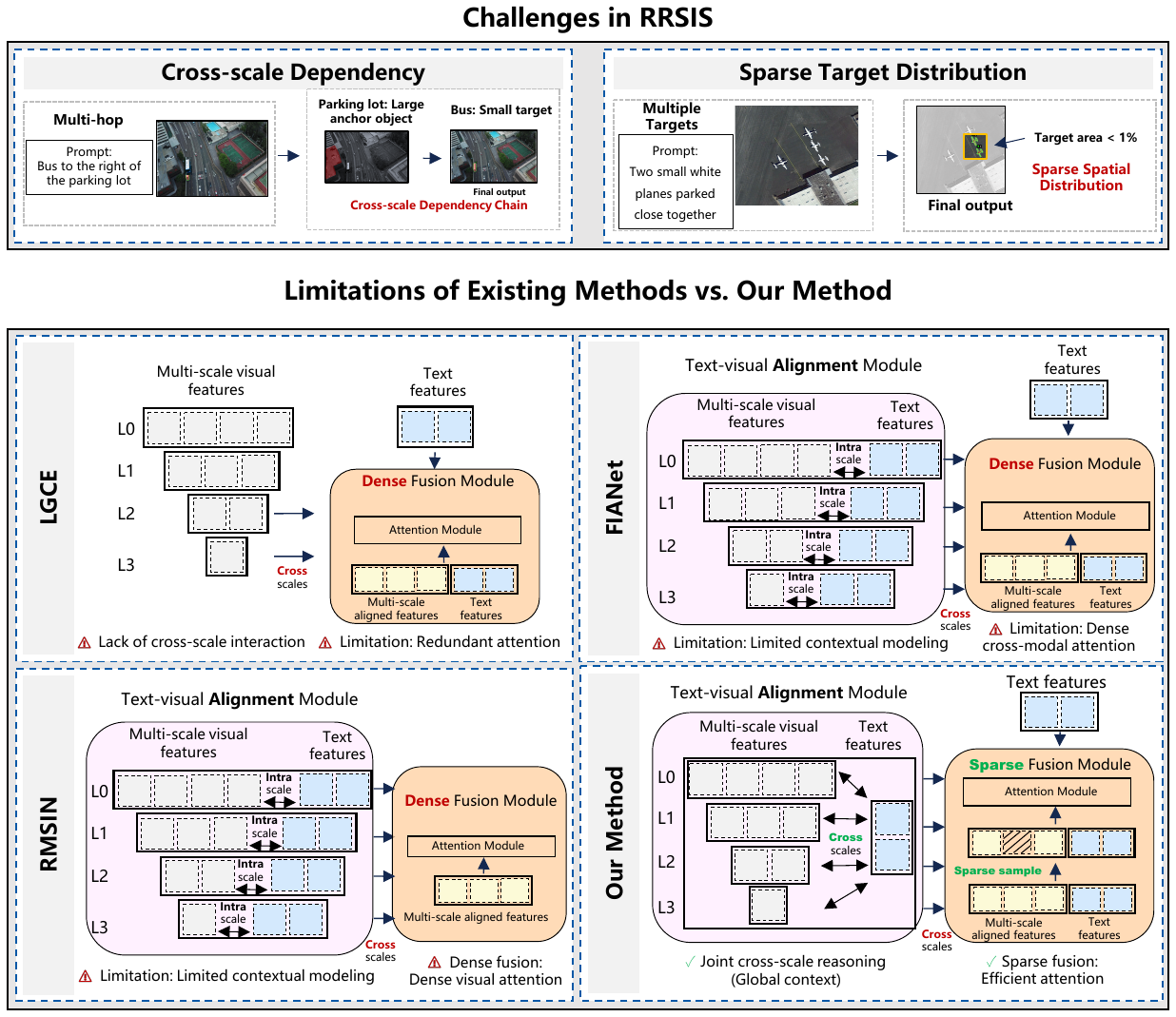}
  \caption{Overview of the motivation and model design. Top: cross-scale dependency and sparse target distribution in RRSIS. Bottom: schematic comparison of LGCE, FIANet, and RMSIN with CSSAM's cross-scale fusion and sparse attention.}
  \label{fig:Figure_1}
\end{figure}

Recent progress in RRSIS has been made through increasingly sophisticated text-visual fusion strategies. LAVT \cite{6} employed pixel-word attention to directly link words with spatial locations; LGCE \cite{7} used language to guide cross-scale visual enhancement; RMSIN \cite{8} introduced modules for intra- and inter-scale feature refinement; while FIANet \cite{9} parsed referring expressions into structured semantic parts. RSRefSeg \cite{10} leveraged pre-trained models like CLIP and SAM to combine global semantics with segmentation priors. Despite these advances, jointly modeling cross-scale visual context and linguistic dependencies remains important. Moreover, the sparsity and spatial dispersion of relevant targets in high-resolution RS imagery motivate reducing unnecessary dense attention interactions. Fig.~\ref{fig:Figure_1} illustrates these challenges and compares representative LGCE, RMSIN, and FIANet designs with CSSAM.

In addition to architectural challenges, existing RRSIS datasets underrepresent key semantic complexities inherent to RS scenes. Referring expressions often involve (1) multi-target references, where multiple regions satisfy a description, sometimes also requiring the model to possess counting ability (e.g., ``three large baseball fields''); (2) multi-hop expressions in the reference-dependent grounding sense \cite{42}, where localization uses another object (e.g., ``tennis courts below the parking lot''); and (3) multi-granularity descriptions for the same target, from simple nouns (``vehicle'') to complex attribute- and relation-based expressions (``red car between two black cars''). These aspects introduce ambiguity and contextual grounding demands, yet are rarely annotated explicitly in current benchmarks.

To bridge this gap, we introduce the Cross-Scale Sparse Attention Model (CSSAM), a unified architecture integrating cross-scale multimodal fusion and query-dependent sparse attention for remote sensing referring segmentation. The Cross-scale Modality-aligned Fusion Module (CMFM) combines bidirectional cross-modal attention with text self-attention and multi-scale deformable visual self-attention. The Dynamic Multimodal Sparse Hybrid Attention (DMSHA) then predicts sampling offsets from each visual query and the pooled text representation, and attends to the selected visual keys and values together with text tokens. Unlike dense attention, DMSHA limits visual interactions to sampled locations; compared with conventional hierarchical fusion, it explicitly predicts sampling locations across scales based on each visual query and the referring expression. Our main methodological novelty lies in DMSHA's text-conditioned cross-scale sampling.

To support the training and evaluation of models like CSSAM, we construct the M$^{3}$RRSIS-D dataset, which explicitly covers three core complexities of RS referring segmentation: Multi-target descriptions (M$^{1}$), Multi-hop relational reasoning (M$^{2}$), and Multi-granularity expressions for the same target (M$^{3}$). Grounded in diverse RS scenes, M$^{3}$RRSIS-D enables explicit assessment of model performance across multiple reasoning types, providing a benchmark closely aligned with real-world remote sensing applications.

In summary, the main contributions of our work are as follows:

(1) A dedicated dataset: We introduce M$^{3}$RRSIS-D, a benchmark for referring segmentation in remote sensing that systematically covers three key complexities, namely multi-target expressions, reference-dependent multi-hop reasoning, and multi-granularity descriptions of the same target or group of targets.

(2) An integrated model: We propose CSSAM, a Transformer-based architecture designed for cross-scale multimodal alignment in RS scenes. It includes CMFM, which bridges text and multi-scale visual features through cross- and self-attention, enabling consistent reasoning across spatial levels.

(3) A sparse attention module: We design DMSHA, which reduces dense visual interactions by sampling cross-scale visual key-value pairs and attending jointly to these samples and the text tokens. Sampling locations depend on both the visual query and the referring expression.

\section{Related work}
\label{sec:2}

\paragraph{Referring Segmentation in Remote Sensing Imagery}
RRSIS aims to segment image regions based on natural language expressions, requiring fine-grained alignment between high-resolution visual data and linguistic semantics. Early methods for referring segmentation typically use CNNs for visual encoding and RNNs for language processing, with cross-modal embeddings for feature alignment, such as the end-to-end CNN-RNN model by Hu et al. \cite{11} and the Recurrent Refinement Network (RRN) by Li et al. \cite{12}, which progressively refines multiscale features. Attention-based cross-modal methods have also been widely studied. Representative works include CMSA \cite{13} for cross-modal self-attention, LSCM \cite{14} for linguistically structured context modeling, CMPC \cite{15} for progressive entity-relation reasoning, BRINet \cite{16} for bi-directional cross-modal dependency modeling, CMPC+ \cite{17} which is an improved CMPC-based progressive comprehension framework with text-guided multi-level feature refinement, and CrossVLT \cite{18} for mutual early fusion between vision and language encoders across all stages to enhance cross-modal context modeling. LAVT \cite{6} achieves early cross-modal fusion within a vision Transformer.

More recently, transformer-based methods have also advanced referring segmentation in RS imagery. LGCE \cite{7} enhances visual features from last two scales through language guidance, RMSIN \cite{8} addresses scale and rotation challenges via multi-scale interaction and adaptive convolutions, FIANet \cite{9} perform fine-grained text-image alignment through expression decoupling, and RSRefSeg \cite{10} leverages CLIP and SAM for robust multimodal feature activation. Zhang et al. \cite{45} further introduce a hierarchical textual--visual guidance framework that parses complex referring expressions and uses hierarchical textual cues to modulate visual representations. Meanwhile, large vision--language models such as LISA \cite{43} and GeoChat \cite{44} demonstrate strong general referring capabilities, but their reliance on large-scale foundation architectures typically entails substantial computational overhead, which can limit practicality in resource-constrained deployment scenarios.

SBANet \cite{sbanet2025} uses learnable query tokens for bidirectional vision--language alignment and a text-conditioned channel and spatial aggregator for cross-scale interaction. DiffRIS \cite{diffris2026} adapts pre-trained text-to-image diffusion features through a context perception adapter and a progressive cross-modal reasoning decoder. CSSAM instead predicts language-conditioned sampling offsets for each visual query and jointly attends to sampled cross-scale visual features and text tokens.

\paragraph{Sparse Attention Strategies for Long-Sequence Transformers}
Transformers excel across vision and language tasks but suffer from quadratic complexity with input length, limiting scalability for high-resolution or long-sequence data. To address this, sparse attention mechanisms have been proposed, broadly categorized into KV cache compression, blockwise KV management, sampling-based selection, and dynamic hierarchical sparsity.

KV cache compression reduces memory by retaining key tokens or compressing stored states, as in H$_{2}$O \cite{19} for heavy-hitter retention and SnapKV \cite{21} for clustering critical positions. LLM$\times$MapReduce \cite{20} adopts divide-and-conquer long-context processing. Blockwise KV management selectively retains full KV caches based on attention roles, with DuoAttention \cite{22} identifying retrieval-critical heads, Quest \cite{23} introducing query-aware KV loading, and SeerAttention \cite{24} learning block-level sparsity through gated attention. Sampling and hash-based selection reduces attention overhead by estimating token importance, as seen in ClusterKV \cite{25} with semantic clustering, HashAttention \cite{26} using Hamming-space mappings, and MagicPIG \cite{27} applying LSH for efficient sparse attention. Integrated dynamic sparsity methods like NSA \cite{28} combine coarse- and fine-grained selection into a hierarchical framework for efficient long-context training and inference.

These sparse attention designs primarily target unimodal long-text modeling, with limited adaptation to high-resolution RS imagery, where sparse and dispersed targets naturally align with sparse modeling strategies.

In visual segmentation, EntroFormer \cite{entroformer2025} selects high-entropy regions to capture sparse global context and combines it with dense local features. CSSAM instead learns sampling locations conditioned on the visual query and referring expression. CTVIS \cite{ctvis2023} addresses a different dependency: it aligns training with inference through instance-embedding learning and a memory bank for temporal association in online video segmentation. Our focus is spatial, language-guided sampling within a single image, not entropy-based region selection or temporal consistency learning.

\paragraph{Datasets for Large-scale RS Scenes}
Existing remote sensing (RS) datasets support tasks like object detection and segmentation. For object detection, the VisDrone dataset \cite{29} comprises 8,599 drone-captured images with over 540,000 bounding boxes. The DIOR-R dataset \cite{30} offers 23,463 images and over 190,000 instances across 20 categories, and the DOTA v2.0 dataset \cite{31} includes 11,268 images across 18 categories, with over 1.7 million object instances. For segmentation, the LoveDA dataset \cite{32} features 5,987 high-resolution images with 166,768 annotated objects. The Vaihingen and Potsdam datasets \cite{33} contain 33 and 38 image patches at ground sampling distances of 9 cm and 5 cm, respectively, with true orthophotos and digital surface models, while the Inria Aerial Image dataset \cite{34} contains 360 high-resolution images ($5000\times5000$ pixels) over 810 km$^{2}$, designed for urban semantic segmentation. Recent efforts have introduced two datasets for RRSIS: RefSegRS \cite{7} provides annotated expressions for object segmentation in RS imagery; and RRSIS-D \cite{8} offers a larger-scale dataset with diverse spatial scales and orientations.

\section{M$^{3}$RRSIS-D}
\label{sec:3}

\begin{figure}[!htbp]
  \centering
  \includegraphics[width=\textwidth]{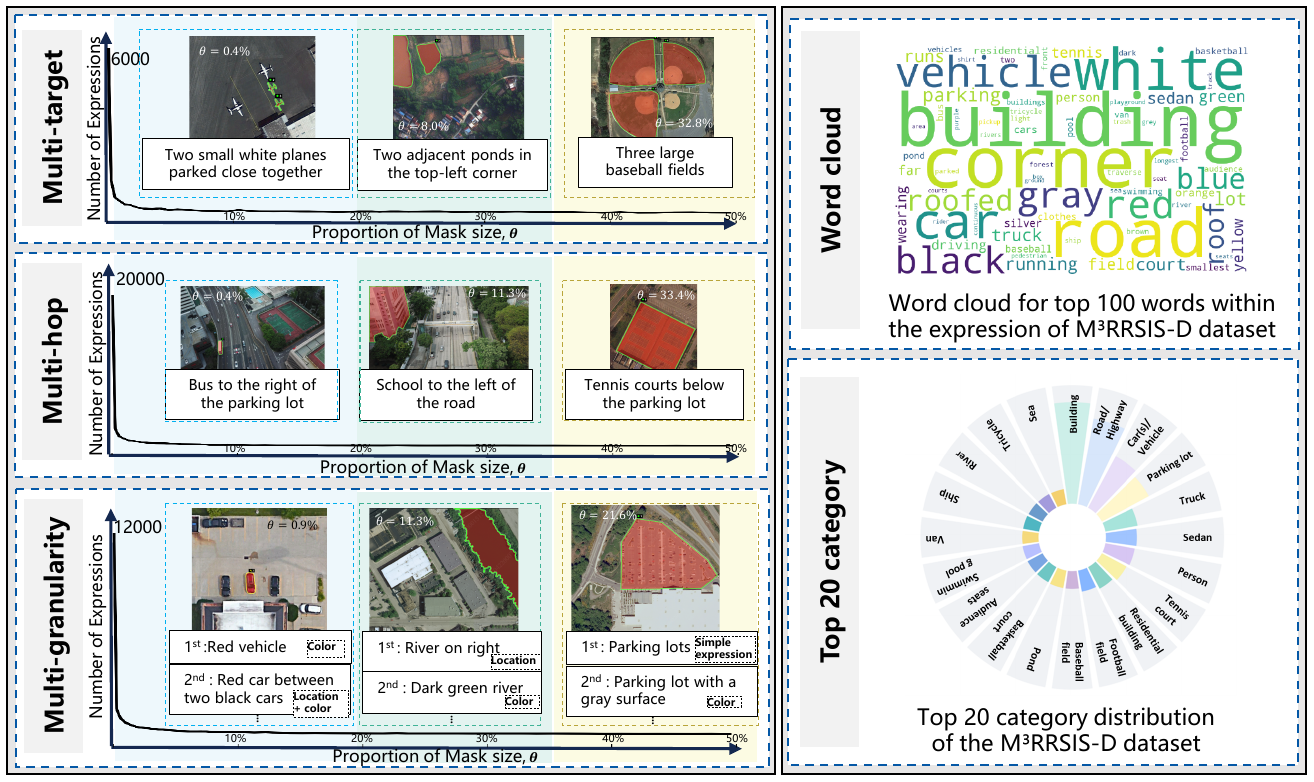}
  \caption{M$^{3}$RRSIS-D dataset. Left: data contribution, showcasing three semantic dimensions in the M$^{3}$RRSIS-D dataset---multi-target (M$^{1}$), multi-hop (M$^{2}$), and multi-granularity (M$^{3}$)---that capture the complexity of real-world remote sensing data. Right: top categories and descriptive keywords of M$^{3}$RRSIS-D.}
  \label{fig:Figure_2}
\end{figure}

This section presents the construction and statistics of M$^{3}$RRSIS-D, a high-resolution RRSIS benchmark comprising training and validation subsets. M$^{3}$RRSIS-D is built from publicly available remote sensing datasets designed for object detection and semantic segmentation, including VisDrone \cite{29}, DIOR-R \cite{30}, DOTA v2.0 \cite{31}, LoveDA \cite{32}, Vaihingen and Potsdam \cite{33}, and Inria Aerial Image \cite{34}. Fig.~\ref{fig:Figure_2} provides an overview of M$^{3}$RRSIS-D, including its three semantic dimensions, word-cloud visualization, and category distribution.

\subsection{Dataset Sources and Annotation Pipeline}
\label{subsec:dataset-sources}

The source datasets were selected to cover diverse RS imaging conditions, spatial resolutions, environments, and object categories. Together, they provide a broad range of aerial scenes with varied object distributions and land-cover characteristics, supporting the construction of diverse referring segmentation samples. The referring expressions and corresponding target polygons were manually annotated and independently verified following the protocol described below.

LoveDA, DIOR-R, DOTA v2.0, and VisDrone images were used without additional processing. Potsdam and Vaihingen images were divided into $1024\times1024$ patches with a fixed stride of 512 pixels in both directions; Inria images were divided into $1000\times1000$ patches with a stride of 500 pixels. No resizing was applied during dataset construction. Annotators selected images supporting at least three referring expressions. For LoveDA, DIOR-R, DOTA v2.0, and VisDrone, the original training and validation assignments of the selected images were preserved. For Potsdam, Vaihingen, and Inria Aerial Image, the resulting image patches were randomly assigned to the training and validation subsets. 

\begin{figure}[!htbp]
  \centering
  \includegraphics[width=\textwidth]{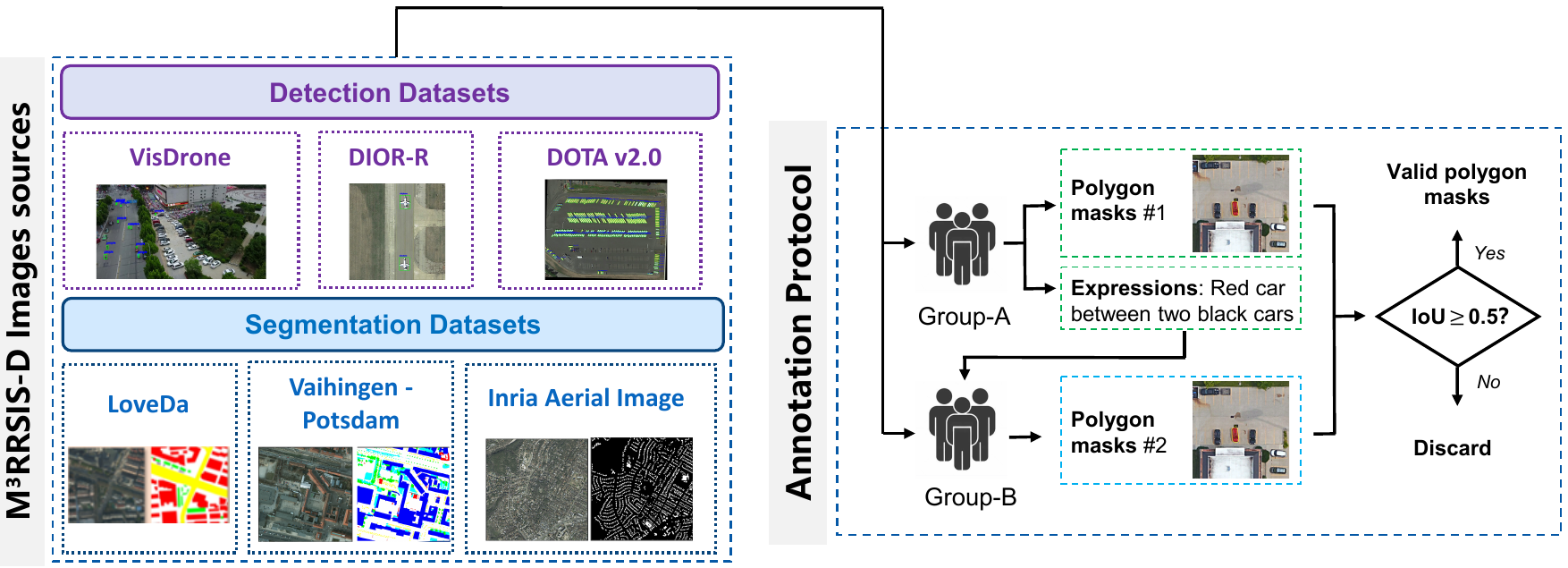}
  \caption{Data sources and annotation workflow of M$^{3}$RRSIS-D.}
  \label{fig:Figure_3}
\end{figure}

Fig.~\ref{fig:Figure_3} illustrates the annotation pipeline. M$^{3}$RRSIS-D adopts a dual-annotator workflow in which Group-A provides the initial polygon masks and referring expressions, while Group-B independently draws masks based solely on the textual description. Only samples satisfying the predefined A/B agreement criterion are retained, ensuring that each retained expression is both spatially grounded and sufficiently unambiguous for independent localization.

\subsection{Annotation Protocol}
\label{subsec:annotation-protocol}

To ensure semantic validity and precise spatial grounding, M$^{3}$RRSIS-D adopts a dual-annotator protocol inspired by ReferItGame \cite{41} and GigaGrounding \cite{42}. The annotation process involves 36 annotators in Group-A and 7 annotators in Group-B and lasted approximately 100 days. The larger Group-A is responsible for the more labor-intensive authoring stage, including image inspection, referable target selection, expression writing, and polygon delineation. Group-B serves as an independent verification group and performs text-only localization: it receives only the referring expression and independently annotates the target polygon without access to the Group-A mask.

\paragraph{Dual-Annotator Workflow}
Group-A generates natural language expressions and delineates corresponding polygon masks. Expressions are categorized as: (i) M$^{1}$ multi-target, describing multiple target instances; (ii) M$^{2}$ multi-hop, specifying targets through relations to other reference objects; and (iii) M$^{3}$ multi-granularity, where up to five hierarchical expressions (L$^{1}$--L$^{5}$) incrementally add attributes such as color, position, object type, shape, or relational context. The labels describe different properties and may overlap.

Group-B independently annotates the target via polygons using only the textual descriptions. An expression is retained only if each referred target satisfies an A/B polygon IoU of at least 0.5. For multi-target cases, both groups must identify the same number of instances; any missing or extra instance is treated as a failed annotation, and every matched instance must satisfy the IoU threshold. Failed samples are revised and re-evaluated; persistently ambiguous or inconsistent cases are discarded.

For M$^{3}$ multi-granularity annotations, each L$^{1}$--L$^{5}$ expression is treated as an independent referring expression and must pass the text-only localization check separately. Therefore, the retained L$^{1}$--L$^{5}$ descriptions are not merely progressively extended attributes; each level is independently valid, unambiguous, and able to localize the corresponding mask. 

During quality control, approximately half of the candidate samples were discarded, mainly due to ambiguous expressions, inconsistent A/B localization, missing or extra instances in multi-target annotations, or masks that failed the IoU agreement criterion. No LLM was used to generate the final referring expressions or polygon annotations.

\paragraph{Semantic Structure of Annotations}
M$^{1}$ (multi-target) expressions refer to more than one target instance to be segmented, e.g., ``three large baseball fields''. Reference objects used only for localization do not count as segmentation targets. Group-A annotates each target instance separately, and Group-B must correctly identify all targets solely from language. This setting evaluates instance grouping and, where specified, numerical cues.

M$^{2}$ (multi-hop) adopts the definition of multi-hop expressions in GigaGrounding \cite{42}, where a target is identified through its relationship with other reference objects. In our dataset, these relationships are spatial, as illustrated by “the school to the right of the river.” This category includes both single target–reference relations and more complex relational chains involving multiple objects.

M$^{3}$ (multi-granularity) labels expressions within annotated groups that share the same target mask but differ in descriptive detail. Up to five levels are provided: L$^{1}$ includes a single salient attribute; L$^{2}$ combines two attributes; L$^{3}$ adds object-type or shape specificity; L$^{4}$ incorporates relational context; and L$^{5}$ integrates multi-object spatial reasoning. Each level is validated independently. The mask may contain one or multiple targets. For example, ``three red houses to the right of the river'' satisfies both the multi-target and relational rules, and also belongs to M$^{3}$ if it is part of such an annotated description group.

\subsection{Dataset Statistics}
\label{subsec:dataset-statistics}

The M$^{3}$RRSIS-D dataset contains 5,603 high-resolution RS images (4,842 for training and 761 for validation) and 66,305 annotated referring expressions (57,720 for training and 8,585 for validation), covering multi-instance targets, relational reasoning, and multi-level semantic granularity.

Across the three semantic dimensions, M$^{1}$ includes 7,723 expressions, M$^{2}$ includes 23,357 expressions, and M$^{3}$ includes 51,976 expressions. Within M$^{3}$, five hierarchical levels (L$^{1}$--L$^{5}$) contain 19,554, 19,554, 9,732, 2,538, and 598 expressions, respectively, reflecting a progressive shift from coarse global semantics to fine-grained local detail. These dimensions are complementary rather than mutually exclusive, and a single expression may belong to multiple dimensions simultaneously.

The pairwise intersections of M$^{1}$ and M$^{2}$, M$^{1}$ and M$^{3}$, and M$^{2}$ and M$^{3}$ contain 2,260, 4,463, and 17,987 expressions in the training set, respectively, compared with 291, 750, and 2,783 in the validation set. The intersection of all three dimensions contains 1,683 training expressions and 254 validation expressions.

M$^{3}$RRSIS-D associates each image with multiple referring expressions, averaging 11.83 expressions per image. As shown by the curves on the left of Fig.~\ref{fig:Figure_2}, mask areas exhibit a long-tailed distribution---most targets are small ($\leq 1\%$), while a few cover large regions such as industrial complexes or airports---supporting comprehensive evaluation of cross-scale semantic representation and fusion.

\subsection{Distinctive Advantages over Existing RRSIS Datasets}
\label{subsec:dataset-advantages}

M$^{3}$RRSIS-D explicitly annotates three complementary dimensions: (1) multi-target expressions, evaluating segmentation of multiple instances and interpretation of numerical cues; (2) multi-hop expressions, assessing reference-dependent relational grounding; and (3) multi-granularity descriptions of the same target mask. Table~\ref{tab:1} distinguishes explicit annotation from the possible presence of a property in existing datasets.

\begin{table}[!htbp]
  \centering
  \caption{Comparison of RRSIS datasets in terms of dataset scale and semantic annotation structure, where ``$\checkmark$'' indicates the dimension is annotated, ``$\circ$'' indicates the dimension may be present but is not annotated, and ``\ding{55}'' indicates that the dimension is absent.}
  \prWideTableSetup
  \begin{tabular*}{\textwidth}{@{\extracolsep{\fill}}lccccc}
    \toprule
    Dataset & \# Images & \# Expressions & M$^{1}$ & M$^{2}$ & M$^{3}$ \\
    \midrule
    RefSegRS \cite{7} & 4,420 & 4,420 & $\circ$ & $\circ$ & \ding{55} \\
    RRSIS-D \cite{8} & 17,402 & 17,402 & \ding{55} & $\circ$ & \ding{55} \\
    \textbf{M$^{3}$RRSIS-D (Ours)} & 5,603 & 66,305 & $\checkmark$ & $\checkmark$ & $\checkmark$ \\
    \bottomrule
  \end{tabular*}
  \label{tab:1}
\end{table}

\section{Cross-Scale Sparse Attention Model}
\label{sec:method}

\subsection{Model Architecture}
\label{subsec:model-architecture}

\begin{figure}[!htbp]
  \centering
  \includegraphics[width=\textwidth]{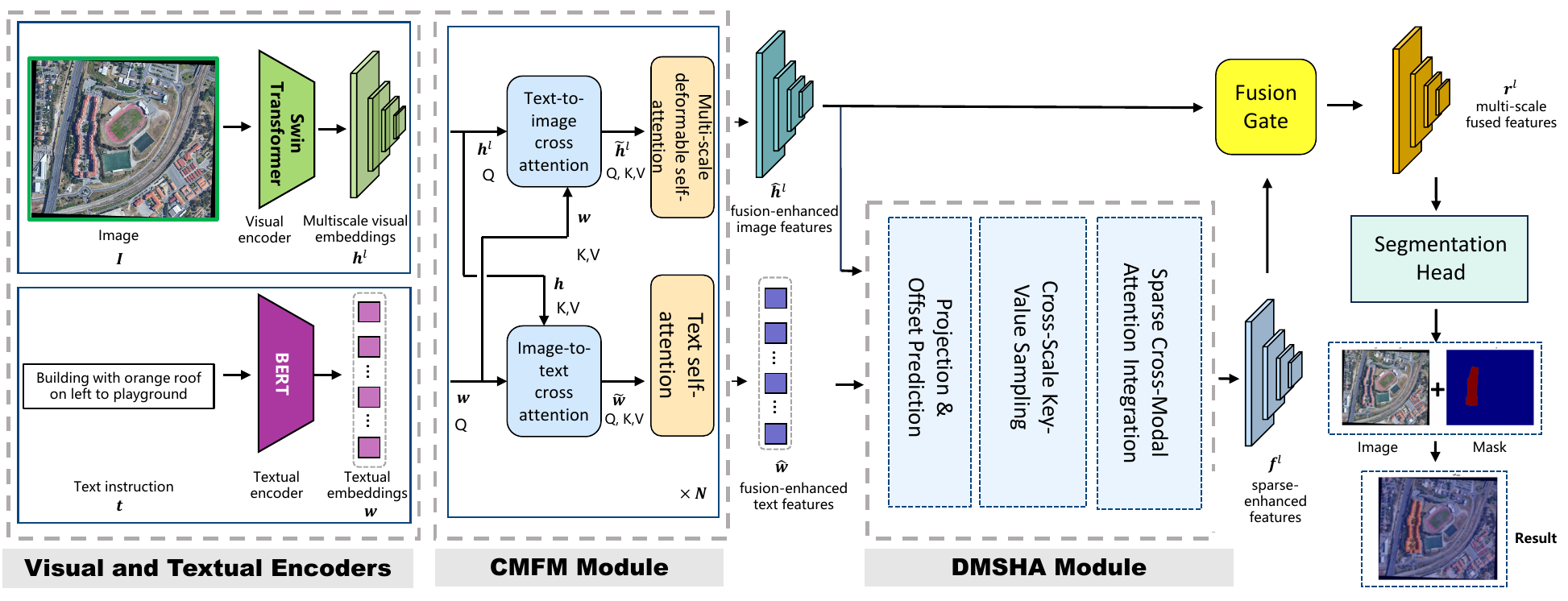}
  \caption{CSSAM architecture: (1) visual and textual encoders, (2) CMFM, (3) DMSHA, (4) a fusion gate, and (5) a segmentation head. The encoders and CMFM attention operations follow established designs; DMSHA introduces query-dependent, text-conditioned cross-scale sampling.}
  \label{fig:Figure_4}
\end{figure}

The overall architecture of CSSAM is depicted in Fig.~\ref{fig:Figure_4}. Given an image $\boldsymbol{I}$ and a query text $\boldsymbol{t}$, CSSAM predicts a binary mask $\boldsymbol{M}$ indicating the region referred to by the query. The model consists of five components: visual and textual encoders that transform the inputs into embeddings; a Cross-scale Modality-aligned Fusion Module (CMFM) that aligns textual and visual features across spatial levels via cross- and self-attention; a Dynamic Multimodal Sparse Hybrid Attention (DMSHA) module that enhances visual features using dynamically sampled multimodal features; a fusion gate that adaptively merges the outputs of CMFM and DMSHA at each scale; and a segmentation head that decodes the final fused features into pixel-level predictions.

\paragraph{Encoder Module}
Given an image-text pair, text embeddings $\boldsymbol{w}=\text{TextEncoder}(\boldsymbol{t})$ are extracted using BERT-base-uncased \cite{37} and projected to $\boldsymbol{w}\in\mathbb{R}^{T\times D}$, where $T$ and $D$ denote the sequence length and embedding dimension, respectively. For the image, Swin-B \cite{38} and the feature projections provide multiscale image embeddings $\boldsymbol{h}^{l}=\text{ImageEncoder}(\boldsymbol{I})$, where $l$ denotes the visual feature level. The embeddings $\boldsymbol{h}^{l}\in\mathbb{R}^{P_l\times D}$, with $P_l=H_l\times W_l$, represent the sequence length at level $l$ and the shared embedding dimension $D$. The implementation uses $L=4$ visual levels and $D=256$.

\subsection{CMFM Module}
\label{subsec:cmfm}

CMFM aligns visual and textual embeddings through cross-modal information exchange. Specifically, image-to-text and text-to-image cross-attention are first applied, followed by modality-specific self-attention. The visual features from all levels are flattened and concatenated as $\boldsymbol{h}=\text{Concat}_{l=0}^{L-1}(\boldsymbol{h}^{l})$, retaining their spatial shapes and level indices. Multi-scale deformable self-attention then aggregates features across these levels.

Formally, given textual embeddings $\boldsymbol{w}$ and the concatenated visual embeddings $\boldsymbol{h}$, cross-modal alignment is performed via
\begin{equation}
    \tilde{\boldsymbol{w}}=\text{CrossAttn}_{\text{img}\to\text{text}}(\text{query}=\boldsymbol{w};\text{key,value}=\boldsymbol{h}),
    \label{eq:cmfm-img2text}
\end{equation}
\begin{equation}
    \tilde{\boldsymbol{h}}^{l}=\text{CrossAttn}_{\text{text}\to\text{img}}(\text{query}=\boldsymbol{h}^{l};\text{key,value}=\boldsymbol{w}),
    \label{eq:cmfm-text2img}
\end{equation}
where each direction updates its query modality using the input features of the same layer. This is followed by intra-modality self-attention:
\begin{equation}
    \begin{aligned}
    \hat{\boldsymbol{w}}&=\text{SelfAttn}(\tilde{\boldsymbol{w}}),\\
    \hat{\boldsymbol{h}}^{l}&=\text{MultiScaleDeformableSelfAttn}
    (\tilde{\boldsymbol{h}}^{l},\{\tilde{\boldsymbol{h}}^{u}\}_{u=0}^{L-1}),
    \end{aligned}
    \label{eq:cmfm-selfattn}
\end{equation}
where $\hat{\boldsymbol{w}}$ and $\hat{\boldsymbol{h}}^{l}$ are the fusion-enhanced text and image features. In the visual self-attention operation, each query at level $l$ samples and aggregates values from all $L$ levels, providing explicit cross-scale exchange rather than independent per-level processing. CMFM contains $N=12$ stacked layers with independent parameters. Each layer contains the cross-modal, text self-attention, and multi-scale visual self-attention operations above, following established multimodal detection and deformable attention designs \cite{39,40}.

\subsection{DMSHA Module}
\label{subsec:dmsha}

\begin{figure}[!htbp]
  \centering
  \includegraphics[width=\textwidth]{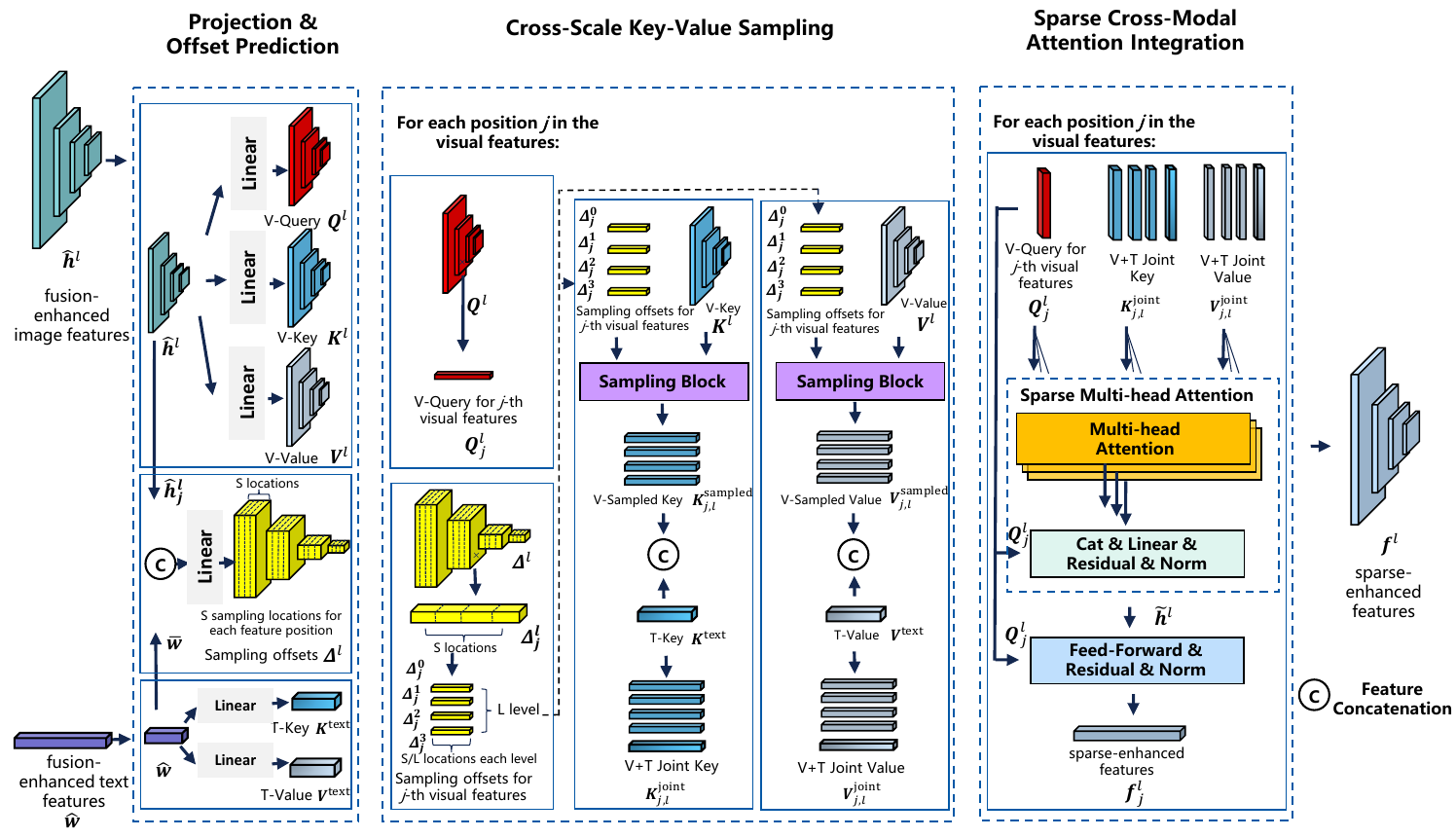}
  \caption{Architecture of the DMSHA module, consisting of three components: (1) projection and offset prediction, (2) cross-scale key--value sampling, and (3) sparse cross-modal attention integration.}
  \label{fig:Figure_5}
\end{figure}

To integrate text features into multiscale visual features without dense visual interactions, we propose DMSHA (Fig.~\ref{fig:Figure_5}). It follows three connected stages: predict where each visual query should sample using the expression, gather visual evidence across scales, and attend jointly to that evidence and the text tokens. DMSHA is applied after the stacked CMFM layers; its output is gated with the CMFM features and decoded into a mask.

\paragraph{Projection and Offset Prediction}
Given fusion-enhanced visual features $\hat{\boldsymbol{h}}^{l}\in\mathbb{R}^{P_l\times D}$ at level $l$, where $P_l=H_l\times W_l$, we apply linear projections to obtain multiscale visual queries, keys, and values:
\begin{equation}
    \boldsymbol{Q}^{l}=\text{Linear}_{Q}(\hat{\boldsymbol{h}}^{l}),\quad
    \boldsymbol{K}^{l}=\text{Linear}_{K}(\hat{\boldsymbol{h}}^{l}),\quad
    \boldsymbol{V}^{l}=\text{Linear}_{V}(\hat{\boldsymbol{h}}^{l}).
    \label{eq:dmsha-qkv}
\end{equation}
Here, $\boldsymbol{Q}^{l}$, $\boldsymbol{K}^{l}$, and $\boldsymbol{V}^{l}\in\mathbb{R}^{P_l\times D}$ correspond to the projected query, key, and value sequences at level $l$. For each spatial position $j$, we extract the visual feature vector $\hat{\boldsymbol{h}}_{j}^{l}\in\mathbb{R}^{D}$ and concatenate it with the averaged text representation $\bar{\boldsymbol{w}}=\text{AvgPool}(\hat{\boldsymbol{w}})\in\mathbb{R}^{D}$. A linear projection then predicts sampling offsets:
\begin{equation}
    \Delta_{j}^{l}=\text{Linear}(\text{Concat}(\hat{\boldsymbol{h}}_{j}^{l},\bar{\boldsymbol{w}}))\in\mathbb{R}^{H\times L\times S\times 2},
    \label{eq:dmsha-offset}
\end{equation}
where $H$ is the number of attention heads, $L$ is the number of visual scale levels, and $S$ is the number of sampled points per scale level. Concatenating the visual and pooled text features produces a vector of length $2D$. A learned affine layer, with weight $\boldsymbol{W}_{\Delta}\in\mathbb{R}^{(2HLS)\times(2D)}$ and bias $\boldsymbol{b}_{\Delta}\in\mathbb{R}^{2HLS}$, maps this vector to $2HLS$ values. These values are reshaped into $H\times L\times S\times2$, specifying horizontal and vertical offsets for each attention head, visual scale, and sampling point. The offsets are defined relative to the normalized reference position $p_j^{\mathrm{ref}}\in[0,1]^2$ of query $j$. Since they are predicted from both the local visual feature and the pooled text feature, the sampling locations depend on the visual query and the referring expression.

Simultaneously, text features $\hat{\boldsymbol{w}}\in\mathbb{R}^{T\times D}$ are projected to text-key and text-value representations:
\begin{equation}
    \boldsymbol{K}^{\text{text}}=\text{Linear}_{\text{text-K}}(\hat{\boldsymbol{w}}),\quad
    \boldsymbol{V}^{\text{text}}=\text{Linear}_{\text{text-V}}(\hat{\boldsymbol{w}}),
    \label{eq:dmsha-textkv}
\end{equation}
with $\boldsymbol{K}^{\text{text}},\boldsymbol{V}^{\text{text}}\in\mathbb{R}^{T\times D}$.

\paragraph{Cross-Scale Key--Value Sampling}
This component allows the model to gather fine-scale details and coarse contextual information without dense visual interactions. Using the predicted offsets $\Delta_j^l$, a sparse set of visual features is sampled from all levels via bilinear interpolation. Let $\mathcal{B}(\cdot,\Delta_j^l)$ denote the sampling operation. The resulting visual key and value tensors for query position $j$ at level $l$ are
\begin{equation}
    \begin{aligned}
    \boldsymbol{K}_{j,l}^{\text{sampled}}&=\mathcal{B}(\{\boldsymbol{K}^{u}\}_{u=0}^{L-1},\Delta_{j}^{l}),\\
    \boldsymbol{V}_{j,l}^{\text{sampled}}&=\mathcal{B}(\{\boldsymbol{V}^{u}\}_{u=0}^{L-1},\Delta_{j}^{l}).
    \end{aligned}
    \label{eq:dmsha-sampled}
\end{equation}
Each head samples $S$ points per level, giving $LS$ visual samples per query. The sampled tensors are reshaped and concatenated with the projected text keys and values to obtain joint cross-modal key and value sets:
\begin{equation}
    \boldsymbol{K}_{j,l}^{\text{joint}}=\text{Concat}(\boldsymbol{K}_{j,l}^{\text{sampled}},\boldsymbol{K}^{\text{text}}),\quad
    \boldsymbol{V}_{j,l}^{\text{joint}}=\text{Concat}(\boldsymbol{V}_{j,l}^{\text{sampled}},\boldsymbol{V}^{\text{text}}).
    \label{eq:dmsha-joint}
\end{equation}

\paragraph{Sparse Cross-Modal Attention Integration}
This component integrates the sampled visual evidence and text tokens according to their relevance to each visual query. For each query position $j$, sparse multi-head attention is performed using query vector $\boldsymbol{Q}_j^l\in\mathbb{R}^{D}$ and joint key/value sets $\boldsymbol{K}_{j,l}^{\text{joint}},\boldsymbol{V}_{j,l}^{\text{joint}}\in\mathbb{R}^{(L\cdot S+T)\times D}$:
\begin{equation}
    \tilde{\boldsymbol{h}}_{j}^{l}=\text{SparseMultiHeadAttn}(\boldsymbol{Q}_{j}^{l},\boldsymbol{K}_{j,l}^{\text{joint}},\boldsymbol{V}_{j,l}^{\text{joint}}).
    \label{eq:dmsha-attn}
\end{equation}
The attention result is refined through a feed-forward network, a residual connection, and layer normalization to produce the final sparse-enhanced features:
\begin{equation}
    \boldsymbol{f}_{j}^{l}=\text{Norm}(\text{FFN}(\tilde{\boldsymbol{h}}_{j}^{l})+\boldsymbol{Q}_{j}^{l}).
    \label{eq:dmsha-output}
\end{equation}

\subsection{Fusion Gate}
\label{subsec:fusion-gate}

Motivated by prior RRSIS works such as LAVT \cite{6} and RMSIN \cite{8}, we incorporate a fusion gate mechanism (Fig.~\ref{fig:Figure_6}a) to adaptively merge the fusion-enhanced image features $\hat{\boldsymbol{h}}^{l}$ and the sparse-enhanced features $\boldsymbol{f}^{l}$ from DMSHA at each scale $l$. Specifically, we define a spatial attention gate $\boldsymbol{g}^{l}$ as
\begin{equation}
    \boldsymbol{g}^{l}=\tanh\left(\text{Conv}_{2}\left(\text{ReLU}\left(\text{Conv}_{1}(\boldsymbol{f}^{l})\right)\right)\right),
    \label{eq:fusion-gate}
\end{equation}
where $\text{Conv}_1$ and $\text{Conv}_2$ are two $1\times1$ convolution layers, ReLU is the rectified linear unit, and $\tanh$ is the hyperbolic tangent function. The final multi-scale fused features are obtained by
\begin{equation}
    \boldsymbol{r}^{l}=\hat{\boldsymbol{h}}^{l}+\boldsymbol{g}^{l}\odot\boldsymbol{f}^{l},
    \label{eq:fusion-output}
\end{equation}
where $\odot$ denotes element-wise multiplication.

\subsection{Segmentation Head}
\label{subsec:segmentation-head}

\begin{figure}[!htbp]
  \centering
  \includegraphics[width=\textwidth]{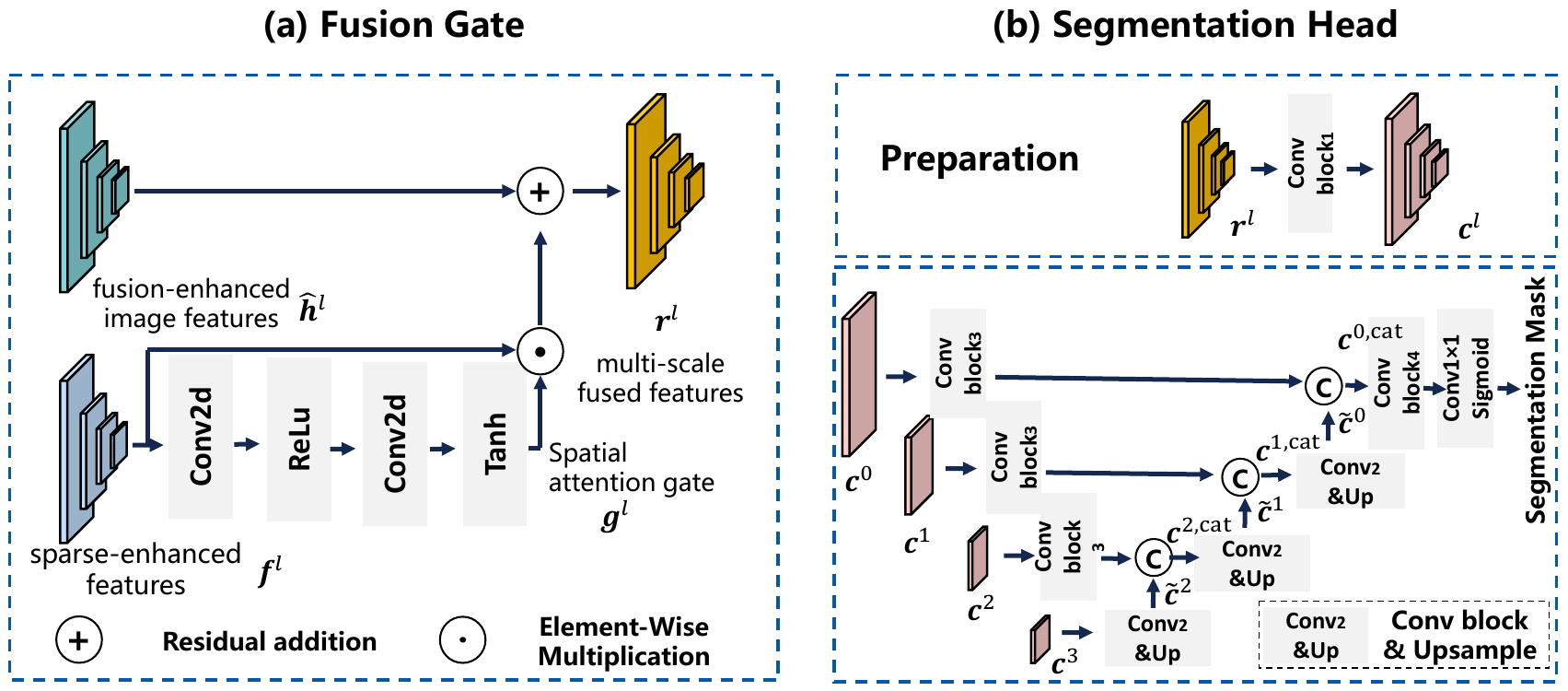}
  \caption{Architecture of (a) the fusion gate and (b) the segmentation head.}
  \label{fig:Figure_6}
\end{figure}

To produce the final binary mask, we adopt a U-Net-like convolutional segmentation head (Fig.~\ref{fig:Figure_6}b) that takes the final multi-scale fused features $\boldsymbol{r}^{l}$ as input. Convolutions refine local features during upsampling. First, each scale is independently processed using a multi-layer convolutional block to adjust the channel dimensions of feature maps:
\begin{equation}
    \boldsymbol{c}^{l}=\text{ConvBlock}_{1}(\boldsymbol{r}^{l}),\quad l=0,1,2,3,
    \label{eq:seg-head-prep}
\end{equation}
where $\text{ConvBlock}_{1}$ consists of three convolutional layers with kernel sizes $1\times1$, $3\times3$, and $1\times1$, respectively, with ReLU activations after the first two.

Starting from the lowest-resolution feature $\boldsymbol{c}^{3}$, each feature is refined and progressively upsampled, then merged with higher-resolution features:
\begin{equation}
    \tilde{\boldsymbol{c}}^{l-1}=\text{Up}(\text{ConvBlock}_{2}(\boldsymbol{c}^{l,\text{cat}})),\quad
    \boldsymbol{c}^{3,\text{cat}}=\boldsymbol{c}^{3},\quad l=3,2,1,
    \label{eq:seg-head-up}
\end{equation}
\begin{equation}
    \boldsymbol{c}^{l-1,\text{cat}}=\text{Concat}(\tilde{\boldsymbol{c}}^{l-1},\text{ConvBlock}_{3}(\boldsymbol{c}^{l-1})),\quad l=3,2,1,
    \label{eq:seg-head-concat}
\end{equation}
where $\text{ConvBlock}_{2}$ refines the merged features without changing channels, while $\text{ConvBlock}_{3}$ processes higher-resolution features before concatenation. This upsample-and-merge process repeats until the highest resolution. The final feature is then passed through a $1\times1$ convolution and a sigmoid function to produce the binary mask:
\begin{equation}
    \boldsymbol{M}=\text{Sigmoid}(\text{Conv}_{1\times1}(\text{ConvBlock}_{4}(\boldsymbol{c}^{0,\text{cat}}))).
    \label{eq:seg-head-mask}
\end{equation}

\subsection{Complexity and Memory Analysis of DMSHA Module}
\label{subsec:complexity}

Let $M=\sum_l P_l$ denote the total number of visual tokens. Holding the head count and feature dimension fixed, dense joint attention has $O(M(M+T))$ attention-score memory, whereas DMSHA uses $O(M(LS+T))$: each visual query attends to $LS$ sampled visual positions and $T$ text tokens. The visual-only interaction term is reduced from $O(M^2)$ to $O(MLS)$. This analysis concerns attention within DMSHA, rather than the cost of backbones, CMFM, or segmentation head.

\subsection{Loss Function}
\label{subsec:loss}

We adopt a hybrid objective that combines cross-entropy loss and Dice loss to jointly optimize pixel-wise classification accuracy and region-level overlap. Specifically, cross-entropy focuses on hard pixel discrimination, while Dice loss directly maximizes mask similarity by measuring the overlap between prediction and ground truth. The final loss is formulated as
\begin{equation}
    \mathcal{L}=(1-\lambda)\mathcal{L}_{\mathrm{CE}}+\lambda\mathcal{L}_{\mathrm{Dice}},
    \label{eq:loss}
\end{equation}
where $\lambda=0.1$ is the default. In the ablation study, we also evaluate $\lambda=0$ and $0.5$. 

\section{Experiments and Evaluations}
\label{sec:experiments}

\subsection{Implementation Details}
\label{subsec:implementation}

\paragraph{Experiment Settings}
CSSAM employs Swin-B \cite{38} (window size 12) and BERT-base-uncased \cite{37} as visual and textual encoders, respectively, initialized from the pretrained Grounding DINO-B checkpoint \cite{39} provided by MMDetection. Both encoders are frozen, while the remaining modules are trained for 20 epochs using AdamW (learning rate $10^{-3}$, weight decay $10^{-4}$) without warm-up. Training uses 10 NVIDIA RTX 4090 GPUs with a total batch size of 20.

Images and masks are resized to $800\times800$ using bilinear and nearest-neighbor interpolation, respectively. Images are normalized using ImageNet statistics, without random cropping, flipping, or color augmentation. Text is tokenized using uncased WordPiece with a maximum length of 256 tokens. CMFM consists of 12 layers with 256-dimensional features, four cross-modal attention heads, and eight deformable visual attention heads over four scales with four sampling points per scale. DMSHA uses 256-dimensional features, eight heads, and the same four-scale sampling configuration.

\paragraph{Evaluation Metrics}
Following prior RRSIS studies \cite{6,7,8,9,10}, we report mean Intersection-over-Union (mIoU), overall Intersection-over-Union (oIoU), and Precision (P@$\tau$), where $\tau\in\{0.5,0.6,0.7,0.8,0.9\}$. mIoU averages the per-query IoUs; oIoU divides the summed intersections by the summed unions. P@$\tau$ is the percentage of queries with IoU $\geq\tau$. Accuracy values are percentages. In comparison tables, bold and underlining indicate the best and second-best values.

\subsection{Comparison with State-of-the-Art Methods on M$^{3}$RRSIS-D}
\label{subsec:sota-m3rrsisd}

\paragraph{Quantitative Results}
To evaluate the effectiveness of CSSAM, we compare it against a comprehensive set of SOTA referring segmentation methods, including LSTM+ResNet101 baselines \cite{15,17} and recent Transformer-based architectures such as LAVT \cite{6}, RMSIN \cite{8}, FIANet \cite{9}, and RSRefSeg \cite{10}. All methods are retrained and evaluated on M$^{3}$RRSIS-D using their original training settings.

As shown in Table~\ref{tab:2}, CSSAM consistently outperforms all competing methods across all metrics. It achieves the highest oIoU and mIoU, outperforming the strongest baselines by 1.60 and 2.79 percentage points, respectively. CSSAM also achieves consistent improvements across all IoU thresholds, reflecting balanced gains in both segmentation accuracy and localization precision. These results demonstrate the overall robustness of the proposed cross-scale fusion and sparse attention design.

\begin{table}[!htbp]
  \centering
  \caption{Comparison of CSSAM with state-of-the-art methods on M$^{3}$RRSIS-D. CMPC/CMPC+ use R-101 and LSTM; LAVT, RMSIN, FIANet, and CSSAM use Swin-B and BERT; RSRefSeg uses CLIP+SAM.}
  \prWideTableSetup
  \setlength{\tabcolsep}{3pt}
  \begin{tabular*}{\textwidth}{@{\extracolsep{\fill}}lrrrrrrr}
    \toprule[1pt]
    Method & P@0.5 & P@0.6 & P@0.7 & P@0.8 & P@0.9 & oIoU & mIoU\\
    \midrule
    CMPC \cite{15} & 64.07 & 57.61 & 49.32 & 37.23 & 19.61 & 71.70 & 57.50\\
    CMPC+ \cite{17} & 64.03 & 57.71 & 49.64 & 37.43 & 20.14 & 71.32 & 57.08\\
    LAVT \cite{6} & 73.35 & 67.83 & 61.02 & 49.99 & 30.03 & 80.42 & 65.93\\
    RMSIN \cite{8} & 63.75 & 58.00 & 51.02 & 41.98 & 27.60 & 76.50 & 58.03\\
    FIANet \cite{9} & \underline{79.65} & \underline{74.16} & \underline{67.45} & \underline{56.11} & 33.55 & 80.77 & \underline{72.63}\\
    RSRefSeg-l \cite{10} & 74.86 & 70.46 & 64.45 & 55.32 & \underline{36.90} & \underline{80.85} & 67.99\\
    \midrule
    \textbf{CSSAM (Ours)} & \textbf{83.86} & \textbf{79.83} & \textbf{74.36} & \textbf{62.73} & \textbf{39.94} & \textbf{82.45} & \textbf{75.42}\\
    \bottomrule[1pt]
  \end{tabular*}
  \label{tab:2}
\end{table}

\begin{figure}[!htbp]
  \centering
  \includegraphics[width=\textwidth]{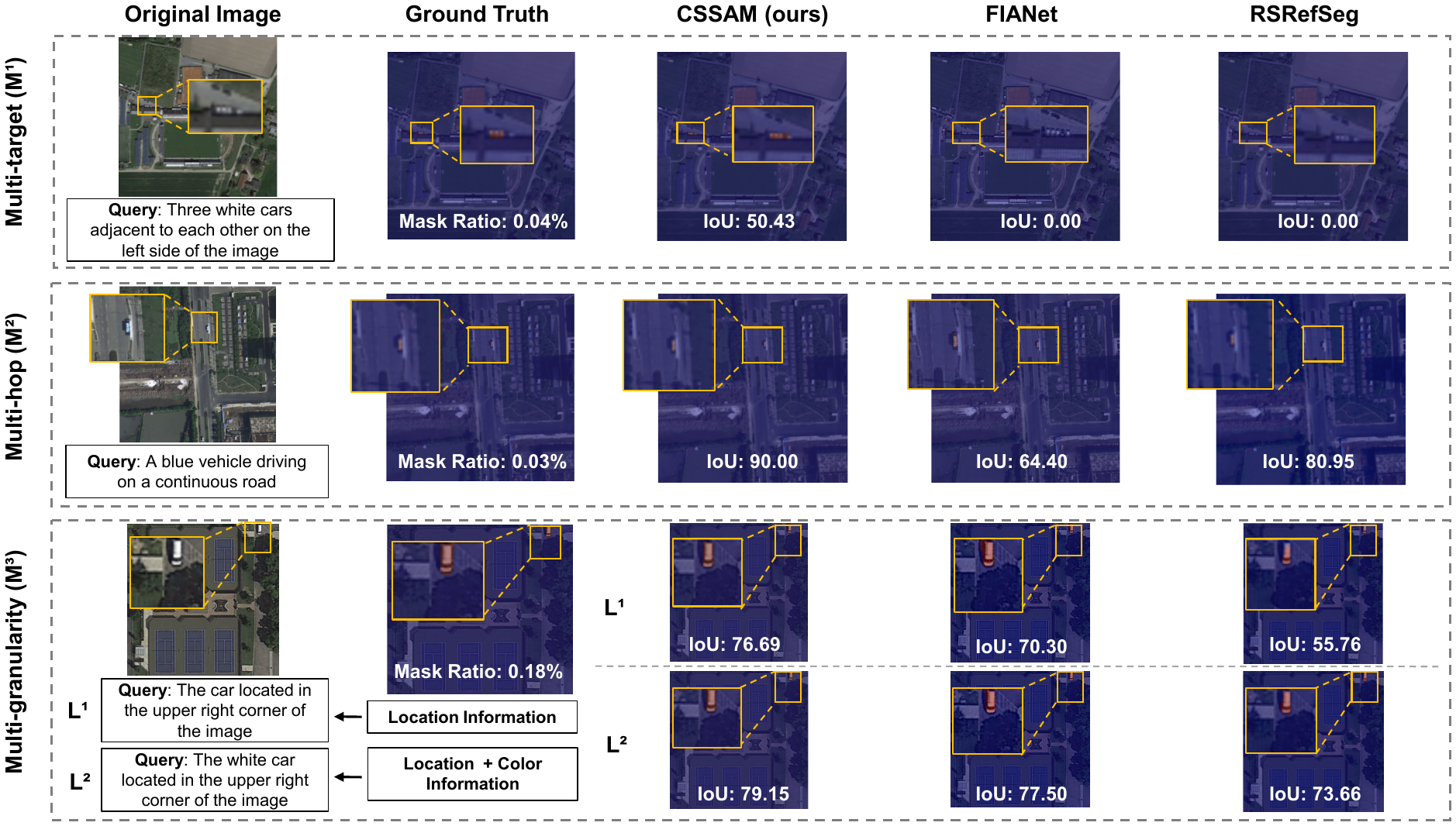}
  \caption{Visualization examples on M$^{1}$ (multi-target), M$^{2}$ (multi-hop reasoning), and M$^{3}$ (multi-granularity) queries.}
  \label{fig:Figure_7}
\end{figure}

\paragraph{Qualitative Results}
Representative visual results of CSSAM, FIANet, and RSRefSeg under multi-target (M$^{1}$), multi-hop (M$^{2}$), and multi-granularity (M$^{3}$) settings are depicted in Fig.~\ref{fig:Figure_7}. These examples illustrate more complete masks for multiple targets and better localization under relational and fine-grained descriptions.

\subsection{Fine-Grained Evaluation by Reasoning Dimension}
\label{subsec:fine-grained-evaluation}

Table~\ref{tab:dimensions}(a) compares CSSAM with LAVT \cite{6}, FIANet \cite{9}, and RSRefSeg \cite{10} on the three reasoning dimensions. CSSAM achieves the highest mIoU and precision at all reported thresholds across the three dimensions. It also achieves the highest oIoU for M$^{2}$ and M$^{3}$, while LAVT leads on M$^{1}$ oIoU.

Table~\ref{tab:granularity}(b) further details the results across granularity levels L$^{1}$--L$^{5}$. CSSAM has the highest mIoU at every level, but not the highest oIoU throughout: LAVT leads on L$^{1}$, L$^{3}$, and L$^{5}$, and RSRefSeg leads on L$^{4}$. FIANet also exceeds CSSAM in L$^{5}$ oIoU. This distinction reflects the different weighting of individual queries and foreground area in the two metrics.

\begin{table}[!htb]
  \centering
  \caption{Fine-grained comparison on M$^{3}$RRSIS-D: (a) the three reasoning dimensions; (b) M$^{3}$ granularity levels, with each cell reporting mIoU/oIoU. Rankings are within each dimension in (a), and within each metric and level in (b).}
  \prWideTableSetup
  \begin{tabular*}{\textwidth}{@{\extracolsep{\fill}}llrrrrr}
    \toprule[1pt]
    \multicolumn{7}{l}{\textbf{(a) Reasoning dimensions}}\\
    \midrule
 Subset & Method & P@0.5 & P@0.7 & P@0.9 & oIoU & mIoU\\
    \midrule
 M$^{1}$ & LAVT \cite{6} & 63.25 & \underline{44.61} & \underline{18.46} & \textbf{76.74} & 57.71\\
    & FIANet \cite{9} & \underline{64.72} & 41.68 & 16.27 & 72.43 & \underline{58.99}\\
    & RSRefSeg-l \cite{10} & 63.13 & 44.59 & 18.14 & 71.63 & 57.10\\
    & \textbf{CSSAM (Ours)} & \textbf{68.24} & \textbf{52.00} & \textbf{19.54} & \underline{74.35} & \textbf{63.94}\\
    \midrule
 M$^{2}$ & LAVT \cite{6} & 65.04 & 52.82 & 25.27 & \underline{78.93} & 58.93\\
    & FIANet \cite{9} & 67.80 & 53.02 & 25.28 & 75.76 & 61.73\\
    & RSRefSeg-l \cite{10} & \underline{69.80} & \underline{59.82} & \underline{33.76} & 78.64 & \underline{63.58}\\
    & \textbf{CSSAM (Ours)} & \textbf{81.37} & \textbf{72.23} & \textbf{37.09} & \textbf{81.58} & \textbf{73.32}\\
    \midrule
    M$^{3}$ & LAVT \cite{6} & 72.82 & 61.58 & 33.28 & \underline{83.24} & 66.39\\
    & FIANet \cite{9} & \underline{76.25} & 62.57 & 31.69 & 80.50 & \underline{68.68}\\
    & RSRefSeg-l \cite{10} & 75.36 & \underline{65.37} & \underline{38.62} & 81.00 & 68.63\\
    & \textbf{CSSAM (Ours)} & \textbf{83.88} & \textbf{75.08} & \textbf{41.48} & \textbf{83.59} & \textbf{75.84}\\
    \midrule[0.8pt]
    \multicolumn{7}{l}{\textbf{(b) Granularity levels: mIoU/oIoU}}\\
    \midrule
    \multicolumn{2}{l}{Method} & L$^{1}$ & L$^{2}$ & L$^{3}$ & L$^{4}$ & L$^{5}$\\
    \midrule
    \multicolumn{2}{l}{LAVT \cite{6}} & 67.40/\textbf{83.33} & 63.40/\underline{82.36} & 68.97/\textbf{85.04} & 71.75/81.76 & \underline{80.93}/\textbf{90.67}\\
    \multicolumn{2}{l}{FIANet \cite{9}} & \underline{69.57}/82.25 & 65.17/77.20 & \underline{71.85}/81.51 & \underline{76.60}/83.36 & 74.94/\underline{87.29}\\
    \multicolumn{2}{l}{RSRefSeg-l \cite{10}} & 68.33/80.77 & \underline{66.05}/79.53 & 70.11/81.66 & 75.35/\textbf{86.50} & 76.25/80.36\\
    \multicolumn{2}{l}{\textbf{CSSAM (Ours)}} & \textbf{75.10}/\underline{82.86} & \textbf{74.64}/\textbf{83.56} & \textbf{76.91}/\underline{84.08} & \textbf{80.76}/\underline{85.22} & \textbf{83.91}/85.16\\
    \bottomrule[1pt]
  \end{tabular*}
  \label{tab:dimensions}
  \label{tab:M1}
  \label{tab:M2}
  \label{tab:M3}
  \label{tab:granularity}
  \label{tab:L1-5-oiou}
  \label{tab:L1-5-miou}
\end{table}

\begin{figure}[!htb]
  \centering
  \includegraphics[width=\textwidth]{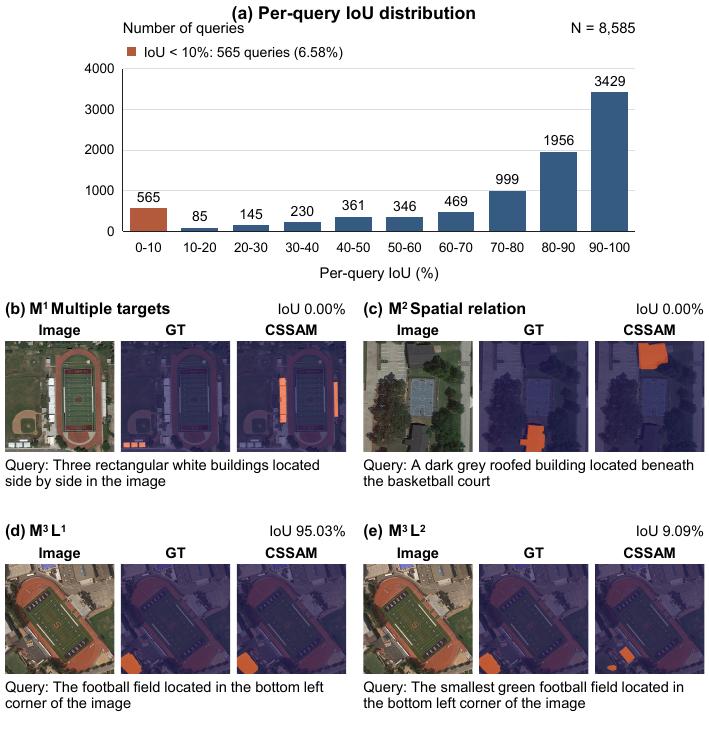}
  \caption{Validation IoU distribution and illustrative failures. (a) Counts for 8,585 queries in left-inclusive, right-exclusive bins; the last bin includes 100\%. (b--c) M$^{1}$ and M$^{2}$ failures. (d--e) M$^{3}$ expressions sharing the same image and ground truth. Each example shows the image, ground truth, and CSSAM prediction.}
  \label{fig:Figure_8}
\end{figure}

\paragraph{Per-query Distribution and Failure Cases}
Across all validation expressions, CSSAM obtains 75.42\% mIoU and 82.45\% oIoU. Figure~\ref{fig:Figure_8}a shows the per-query IoU distribution. The median IoU is 86.30\%, with the 25th and 75th percentiles at 69.03\% and 93.99\%, respectively. The standard deviation is 27.03 percentage points. The 565 queries with IoU below 10\% account for 6.58\% of the split, indicating severe segmentation failures.

Figure~\ref{fig:Figure_8}b--e illustrates three failure cases. In M$^{1}$, the model misses three small white buildings and selects larger white structures near the sports field (IoU 0\%); similar color and shape may obscure the requested group. In M$^{2}$, it selects a building above rather than below the basketball court (IoU 0\%), suggesting confusion between similar roofs when applying the spatial constraint. For M$^{3}$, the same football-field mask obtains 95.03\% IoU with the L$^{1}$ query but only 9.09\% after adding ``smallest green'' at L$^{2}$: most of the field is missed and a nearby basketball court is selected. More descriptive detail therefore does not always yield consistent grounding among adjacent, visually similar facilities.

\subsection{Ablation Studies}
\label{subsec:ablation}

We conduct ablations on the M$^{3}$RRSIS-D validation set to evaluate each component of CSSAM. Table~\ref{tab:ablation-summary} groups the experiments by module design, sampling mode, loss weight, segmentation head, and sampling allocation. The shared default uses CMFM and DMSHA, dynamic sampling with $[4,4,4,4]$ points across scales, loss weight $\lambda=0.1$, and the convolutional segmentation head. The sampling-budget and efficiency results are reported together in Table~\ref{tab:4}.

\paragraph{Effectiveness of CMFM and DMSHA}
The ablation results in Table~\ref{tab:ablation-summary} show that both CMFM and DMSHA improve segmentation performance when introduced individually, while their combination achieves best results. 

\begin{table}[!htb]
  \centering
  \caption{Component and design ablations on M$^{3}$RRSIS-D. \ding{51}/\ding{55} denote enabled/disabled modules. Each ablation is compared with the shared default from Table~\ref{tab:2}. Bold and underlined values mark the best and second best in groups with multiple variants. Internal-design, loss, head, and allocation variants use final epoch-20 weights and all 8,585 validation expressions.}
  \prWideTableSetup
  \setlength{\tabcolsep}{2pt}
  \begin{tabular*}{\textwidth}{@{\extracolsep{\fill}}p{64pt}ccrrrrr@{}}
    \toprule[1pt]
    Category & \multicolumn{2}{c}{Configuration} & P@0.5 & P@0.7 & P@0.9 & oIoU & mIoU\\
    \midrule
    \multirow{7}{64pt}{\textbf{Module design}} & CMFM & DMSHA & \multicolumn{5}{c}{}\\
    & \ding{51} & \ding{51} & \textbf{83.86} & \textbf{74.36} & \textbf{39.94} & \textbf{82.45} & \textbf{75.42}\\
    & \ding{55} & \ding{55} & 62.64 & 46.02 & 18.90 & 69.98 & 57.05\\
    & \ding{51} & \ding{55} & 75.55 & 62.93 & 27.82 & 73.35 & 65.13\\
    & \ding{55} & \ding{51} & \underline{79.46} & \underline{67.52} & \underline{33.94} & \underline{80.76} & \underline{70.95}\\
    \cmidrule(l){2-8}
    & \multicolumn{2}{l}{\makecell[l]{CMFM: identity\\visual attention}} & 76.87 & 64.14 & 30.95 & 79.21 & 68.97\\
    \midrule[0.8pt]
    \multirow{2}{64pt}{\textbf{Sampling strategy}} & \multicolumn{2}{l}{Direct feature} & 79.77 & 67.33 & 31.97 & \underline{80.96} & 71.64\\
    & \multicolumn{2}{l}{Fixed neighbors} & \underline{81.19} & \underline{68.26} & \underline{32.34} & 80.64 & \underline{72.06}\\
    \midrule[0.8pt]
    \multirow{2}{64pt}{\textbf{Loss weight}} & \multicolumn{2}{l}{$\lambda=0$ (CE only)} & 80.45 & \underline{70.08} & 35.96 & 80.45 & \underline{72.80}\\
    & \multicolumn{2}{l}{$\lambda=0.5$} & \underline{80.56} & 70.06 & \underline{36.72} & \underline{81.25} & 72.78\\
    \midrule[0.8pt]
    \textbf{Segmentation head} & \multicolumn{2}{l}{\makecell[l]{Windowed\\Transformer}} & 80.70 & 70.86 & 36.92 & 80.59 & 72.72\\
    \midrule[0.8pt]
    \multirow{4}{64pt}{\textbf{Sampling allocation}} & \multicolumn{2}{l}{$[8,4,2,2]$} & \underline{81.83} & \underline{71.49} & \underline{38.28} & \underline{81.82} & \underline{74.04}\\
    & \multicolumn{2}{l}{$[4,8,2,2]$} & 81.40 & 70.96 & 36.67 & 80.96 & 73.44\\
    & \multicolumn{2}{l}{$[4,2,8,2]$} & 81.29 & 71.19 & 37.69 & 81.19 & 73.49\\
    & \multicolumn{2}{l}{$[4,2,2,8]$} & 81.49 & 71.29 & 38.15 & 81.28 & 73.55\\
    \bottomrule[1pt]
  \end{tabular*}
  \label{tab:ablation-summary}
  \label{tab:3}
  \label{tab:additional-ablation}
  \label{tab:5}
\end{table}

\paragraph{Internal Design Choices}
Replacing CMFM's deformable visual attention with an identity mapping, while retaining the other operations, reduces mIoU from 75.42\% to 68.97\%. Setting $\lambda=0$ (cross-entropy only) or $\lambda=0.5$ (equal weights for cross-entropy and Dice) yields 72.80\% and 72.78\% mIoU, respectively, compared with 75.42\% at $\lambda=0.1$. Four non-uniform sampling allocations, each using 16 points per head, achieve 73.44\%--74.04\% mIoU, below the uniform allocation (75.42\%). The allocation entries are ordered from highest to lowest visual resolution.

\paragraph{Segmentation Head}
We replace the convolutional blocks with Transformer blocks using four-head self-attention in non-overlapping $8\times8$ windows without window shifting \cite{38}, and a two-layer GELU feed-forward network. The remaining decoder structure is unchanged. Both variants are trained for 20 epochs. The Transformer head achieves 72.72\% mIoU and 80.59\% oIoU, compared with 75.42\% and 82.45\% for the convolutional head.

\paragraph{Sampling Budget and Strategy}
We vary the number of sampled key--value pairs per visual level, $S\in\{2,4,6,8,10\}$, corresponding to $4S$ points per attention head. Table~\ref{tab:4} combines segmentation accuracy with inference memory and speed. Efficiency is profiled on all 8,585 validation expressions with batch size 1, fixed $800\times800$ inputs, and one RTX 4090. 

\begin{table}[!htbp]
  \centering
  \caption{Sampling-budget accuracy and inference efficiency. Memory is in GB, latency in ms/query, and throughput in queries/s. Arrows indicate the preferred direction. OOM indicates out-of-memory.}
  \prWideTableSetup
  \setlength{\tabcolsep}{1.1pt}
  \begin{tabular*}{\textwidth}{@{\extracolsep{\fill}}crrrrrrrrr}
    \toprule[1pt]
    $S$ & P@0.5 & P@0.7 & P@0.9 & oIoU & mIoU & \makecell{Peak\\mem.$\downarrow$} & \makecell{Avg.\\mem.$\downarrow$} & \makecell{Latency\\$\downarrow$} & \makecell{Through-\\put$\uparrow$}\\
    \midrule
 2 & 80.29 & 64.33 & 32.20 & 80.00 & 70.82 & \textbf{14.30} & \textbf{14.09} & \textbf{127.86} & \textbf{7.82}\\
 4 & 83.86 & 74.36 & 39.94 & 82.45 & 75.42 & \underline{17.25} & \underline{16.95} & \underline{130.42} & \underline{7.67}\\
 6 & 84.70 & 74.43 & 40.32 & 82.95 & 75.99 & 21.98 & 21.37 & 134.14 & 7.46\\
 8 & \underline{84.92} & \underline{75.87} & \underline{40.39} & \underline{83.02} & \underline{76.03} & 23.51 & 23.33 & 157.84 & 6.34\\
 10 & \textbf{84.95} & \textbf{75.92} & \textbf{40.40} & \textbf{83.08} & \textbf{76.09} & 23.55 & 23.47 & 164.32 & 6.09\\
    \midrule
    \multicolumn{2}{l}{Dense counterpart} & \multicolumn{4}{c}{\rule[0.5ex]{3.5cm}{0.3pt}} & OOM & \multicolumn{3}{c}{\rule[0.5ex]{2.5cm}{0.3pt}}\\
    \bottomrule[1pt]
  \end{tabular*}
  \label{tab:4}
\end{table}

Increasing $S$ from 4 to 10 improves mIoU by 0.67 points, while peak memory rises from 17.25 to 23.55 GB and throughput falls from 7.67 to 6.09 queries/s. The $S=8$ and $S=10$ runs nearly exhaust the 24 GB device. We therefore use $S=4$ to balance accuracy, memory demand, and speed.

To isolate the role of sparsification, we replace DMSHA's sampled visual attention with dense attention over all cross-scale visual tokens, leaving the other components unchanged. Using the profiling setup of Table~\ref{tab:4}, the dense counterpart runs out of memory, whereas CSSAM completes inference for all tested sampling settings. This supports the memory benefit of sparsification within DMSHA.

The sampling strategy group in Table~\ref{tab:ablation-summary} compares direct feature reuse without spatial sampling, fixed four-neighbor sampling around each reference point, and dynamic sampling that uses a linear layer to predict sampling offsets from visual features and the averaged text embedding. Dynamic sampling achieves the highest scores.

\subsection{Additional Evaluation on Public RRSIS Benchmarks}
\label{subsec:public-benchmarks}

To further evaluate the generalization ability of CSSAM beyond M$^{3}$RRSIS-D, we additionally conduct experiments on two public referring remote sensing image segmentation datasets, RRSIS-D \cite{8} and RefSegRS \cite{7}. CSSAM is trained on the official training split of each dataset and evaluated on the corresponding test split. Following prior RRSIS works, we report P@0.5--P@0.9, oIoU, and mIoU. Results of previous methods are collected from the corresponding public literature, and only methods with available reported results are included. Tables~\ref{tab:rrsisd-public-benchmark} and \ref{tab:refsegrs-public-benchmark} give the results on RRSIS-D and RefSegRS, respectively.

\begin{table}[!htbp]
  \centering
  \caption{Comparison on the RRSIS-D test split. Competing results are literature-reported; ``--'' denotes an unreported metric.}
  \prWideTableSetup
  \setlength{\tabcolsep}{3pt}
  \begin{tabular*}{\textwidth}{@{\extracolsep{\fill}}lrrrrrrr}
    \toprule[1pt]
 Method & P@0.5 & P@0.6 & P@0.7 & P@0.8 & P@0.9 & oIoU & mIoU\\
    \midrule
    CMPC \cite{15} & 55.83 & 47.40 & 36.94 & 25.45 & 9.19 & 69.22 & 49.24\\
    CMPC+ \cite{17} & 57.65 & 47.51 & 36.97 & 24.33 & 7.78 & 68.64 & 50.24\\
    LAVT \cite{6} & 66.93 & 60.99 & 51.71 & 39.79 & 23.99 & 76.58 & 59.05\\
    RMSIN \cite{8} & 71.96 & 65.76 & 55.16 & 42.03 & 25.02 & 76.50 & 62.27\\
    FIANet \cite{9} & 74.46 & 66.96 & 56.31 & 42.83 & 24.13 & 76.91 & 64.01\\
    RSRefSeg-l \cite{10} & 74.49 & 68.33 & \underline{58.73} & \underline{48.50} & \underline{30.80} & 77.24 & 64.67\\
    HTVG \cite{45} & 75.21 & \underline{69.03} & 57.45 & 43.92 & 24.96 & 77.85 & 64.82\\
    SBANet \cite{sbanet2025} & \underline{75.91} & -- & 57.05 & -- & 25.38 & \textbf{79.22} & \underline{65.52}\\
    DiffRIS \cite{diffris2026} & 71.79 & -- & 54.64 & -- & 26.46 & 76.05 & 62.19\\
    \midrule
    \textbf{CSSAM (Ours)} & \textbf{79.26} & \textbf{75.18} & \textbf{68.97} & \textbf{57.25} & \textbf{33.78} & \underline{77.90} & \textbf{70.97}\\
    \bottomrule[1pt]
  \end{tabular*}
  \label{tab:rrsisd-public-benchmark}
\end{table}

\begin{table}[!htbp]
  \centering
  \caption{Comparison on the RefSegRS test split. Competing results are literature-reported; ``--'' denotes an unreported metric.}
  \prWideTableSetup
  \setlength{\tabcolsep}{3pt}
  \begin{tabular*}{\textwidth}{@{\extracolsep{\fill}}lrrrrrrr}
    \toprule[1pt]
 Method & P@0.5 & P@0.6 & P@0.7 & P@0.8 & P@0.9 & oIoU & mIoU\\
    \midrule
    LAVT \cite{6} & 70.23 & 55.53 & 30.05 & 14.42 & 4.07 & 76.21 & 57.30\\
    RMSIN \cite{8} & 71.60 & 55.97 & 31.87 & 11.72 & 1.93 & 71.73 & 57.78\\
    FIANet \cite{9} & \textbf{84.09} & \underline{77.05} & 61.86 & \underline{33.41} & 7.10 & 78.32 & 68.67\\
    HTVG \cite{45} & \underline{83.60} & 76.06 & \underline{63.02} & 32.58 & 7.26 & \textbf{80.12} & \underline{68.81}\\
    SBANet \cite{sbanet2025} & 77.02 & -- & 44.15 & -- & \underline{8.97} & \underline{79.86} & 62.73\\
    DiffRIS \cite{diffris2026} & 74.57 & 64.56 & 46.51 & 24.49 & 5.78 & 76.18 & 62.38\\
    \midrule
    \textbf{CSSAM (Ours)} & 83.49 & \textbf{78.21} & \textbf{67.80} & \textbf{51.51} & \textbf{24.99} & 79.17 & \textbf{73.24}\\
    \bottomrule[1pt]
  \end{tabular*}
  \label{tab:refsegrs-public-benchmark}
\end{table}

The strong performance on both public benchmarks demonstrates that CSSAM generalizes well beyond the newly introduced M$^{3}$RRSIS-D benchmark. On RRSIS-D, CSSAM outperforms SBANet by 5.45 points in mIoU and obtains the highest score on every reported metric except oIoU, where it ranks second. On RefSegRS, FIANet and HTVG are marginally better at P@0.5 and oIoU, respectively, whereas CSSAM improves mIoU by 4.43 points over HTVG and achieves substantially higher P@0.8 and P@0.9. These results indicate that the proposed cross-scale sparse attention design transfers well to existing RRSIS datasets and provides more accurate localization under strict overlap criteria.

\section{Conclusion}
\label{sec:conclusion}

This paper presented CSSAM, a cross-scale sparse attention framework for referring segmentation in remote sensing imagery, together with M$^{3}$RRSIS-D, a benchmark covering multi-target, multi-hop relational,
and multi-granularity expressions. CSSAM combines cross-scale multimodal alignment with dynamically sampled sparse attention, enabling effective interaction between linguistic semantics and visual features at different spatial resolutions. Experiments on M$^{3}$RRSIS-D and two public RRSIS benchmarks demonstrate competitive segmentation performance, particularly under strict IoU thresholds. The proposed benchmark also provides a structured basis for evaluating fine-grained reasoning in future RRSIS studies.

The current study is limited by the geographic and sensor diversity of the source imagery. Failure cases also show missed small target groups, incorrect spatial relations, and inconsistent grounding across expression granularities. Future work will extend the benchmark to broader multi-sensor scenarios, and explore lighter implementations for resource-constrained deployment.

\section*{Acknowledgements}
This work was supported by the Young Scientists Fund-Type C of the National Natural Science Foundation of China (Grant No. 62502048), the Qiyuan National Laboratory Innovation Fund (Grant No. 2025-JCJ0-LA-001-089), and the Research Start-up Fund for Introduced Talents of Beijing University of Posts and Telecommunications (Grant No. 2025KYQD10).

\section*{CRediT authorship contribution statement}
\textbf{Yuxi Li:} Methodology, Writing - original draft, Writing - review \& editing, Software, Data curation. 
\textbf{Lu Si:} Methodology, Writing - original draft, Writing - review \& editing, Software, Data curation. 
\textbf{Xia Liao:} Supervision, Conceptualization, Resources, Project administration, Writing - review \& editing. 
\textbf{Yidong Shen:} Software, Validation, Data curation. 
\textbf{Zheng Wei:} Software, Validation, Data curation. 
\textbf{Jun Zhang:} Supervision, Resources, Writing -- review \& editing. 
\textbf{Xin Yuan:} Conceptualization, Supervision, Resources, Writing -- review \& editing.

\section*{Declaration of competing interest}
The authors declare that they have no known competing financial interests or personal relationships that could have appeared to influence the work reported in this paper.

\section*{Declaration of generative AI and AI-assisted technologies in the manuscript preparation process}
During the preparation and revision of this work, the authors used generative AI tools for spelling and grammar checking, language polishing, and readability improvement. The tools were not used to generate research ideas, data, experimental results. After using the tools, the authors reviewed and edited the content.

\section*{Data and Code Availability}
The M$^3$RRSIS-D dataset, source code, and trained model weights will be made publicly available upon acceptance of the manuscript at \url{https://github.com/lyxyuren/CSSAM}. The repository is currently private and will be made public upon acceptance.

\bibliographystyle{elsarticle-num}
\bibliography{main}

\end{document}